# Electrocardiogram (ECG) Based Cardiac Arrhythmia Detection and Classification using Machine Learning Algorithms

Atit Pokharel*
*Department of Electrical and Electtionics Engineering*
*Kathmandu University*
*Dhulikhel, Nepal*

Shashank Dahal
*Department of Electrical and Electtionics Engineering*
*Kathmandu University*
*Dhulikhel, Nepal*

Pratik Sapkota
*Department of Electrical and Electtionics Engineering*
*Kathmandu University*
*Dhulikhel, Nepal*

Prof. Dr. Bhupendra Bimal Chhetri
*Department of Electrical and Electtionics Engineering*
*Kathmandu University*
*Dhulikhel, Nepal*

*Abstract*— The rapid advancements in Artificial Intelligence, specifically Machine Learning (ML) and Deep Learning (DL), have opened new prospects in medical sciences for improved diagnosis, prognosis, and treatment of severe health conditions. This paper focuses on the development of an ML model with high predictive accuracy to classify arrhythmic electrocardiogram (ECG) signals. The ECG signals datasets utilized in this study were sourced from the PhysioNet and MIT-BIH databases. The research commenced with binary classification, where an optimized Bidirectional Long Short-Term Memory (Bi-LSTM) model yielded excellent results in differentiating normal and atrial fibrillation signals. A pivotal aspect of this research was a survey among medical professionals, which not only validated the practicality of AI-based ECG classifiers but also identified areas for improvement, including accuracy and the inclusion of more arrhythmia types. These insights drove the development of an advanced Convolutional Neural Network (CNN) system capable of classifying five different types of ECG signals with better accuracy and precision. The CNN model's robust performance was ensured through rigorous stratified 5-fold cross-validation. A web portal was also developed to demonstrate real-world utility, offering access to the trained model for real-time classification. This study highlights the potential applications of such models in remote health monitoring, predictive healthcare, assistive diagnostic tools, and simulated environments for educational training and interdisciplinary collaboration between data scientists and medical personnel.

I. INTRODUCTION

Cardiovascular diseases (CVDs) stand as one of the prominent contributors to global morbidity, presenting a significant health challenge worldwide [1], [2]. The onset of potentially fatal cardiac conditions is often marked by arrhythmias, making them critical early warning indicators. Developing and implementing a robust predictive model capable of accurately detecting these arrhythmias could profoundly impact health outcomes. Through early detection and intervention, it holds the potential to substantially mitigate the casualty rates associated with heart diseases [1]–[4].

An Electrocardiogram (ECG) visualizes the heart's electrical activity, with key components being the P, Q, R, S, and T waves, as shown in Figure [1]. The P wave signifies atrial contraction, the QRS complex represents ventricular contraction, and the T wave corresponds to ventricular relaxation [5].

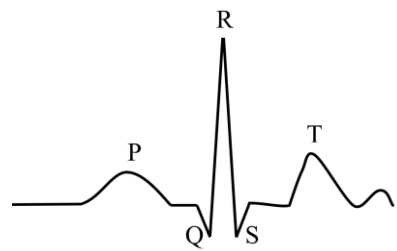

Fig. 1. Basic Components of an ECG Signal

The most critical ECG signal parts and parameters for diagnosing cardiac conditions such as arrhythmias are the QRS complex, ST segment, and PR, RR, and QT intervals. An arrhythmia, essentially an irregular heartbeat rhythm, results to the changes in these ECG waveforms, with atrial fibrillation (Af) being one of the most common types [6], [7]. Numerous techniques can be used for the exploration of ECG signals, such as time-domain-based techniques, frequency domain-based techniques, time frequency-based analysis (TFA), time-fractional frequency domain techniques, and nonlinear [8]–[10]. Multiple research efforts have been conducted to apply machine learning models for diagnosing cardiac arrhythmias using ECG signals with promising results. Ebrahimi et al. [11] review advanced machine learning models essential in medical diagnosis and summarize notable deep learning methods such as Convolutional Neural Networks (CNN), Recurrent Neural Networks (RNN), Deep Belief Networks (DBF) and Long Short-Term Memory Networks (LSTM) for ECG classification. Their analysis of 75 reports from 2017 and 2018 highlights the prominence of CNNs in feature extraction, which was used in 52% of the investigations. The relevance of this study resides in its potential to increase the knowledge and implementation of DL approaches in ECG signal processing, ultimately leading to more accurate and faster cardiac care diagnosis. Another study conducted by Saadatnejad et al. [12] proposed a novel ECG classification algorithm tailored for wearable devices with limited processing capacity. By integrating wavelet transform with multiple LSTM recurrent neural networks, the researchers presented an architecture optimized for continuous cardiac monitoring. Their research illuminates the potential of lightweight LSTM-based ECG classification methods in enhancing continuous monitoring on wearable devices. Another recent study [13] introduced an innovative IoT-based system for heart monitoring and arrhythmia detection using machine learning. The k-nearest neighbor algorithm stood out for its precision, accurately identifying specific types of arrhythmias such as premature ventricular contraction, fusion of ventricular beat, and supraventricular premature beat. This research highlights the transformative potential of combining IoT and machine learning in healthcare, paving the way for enhanced remote patient monitoring and early detection of diverse cardiac irregularities. Support Vector Machine (SVM) is also widely used for ECG classification algorithms. Dhyani et al. [14] present a method combining the 3D Discrete Wavelet Transform (DWT) and Support Vector Machine (SVM) for ECG signal analysis. Using the China Physiological Signal Challenge (CPSC) 2018 dataset, the SVM classifier categorized ECG signals into heartbeat types, achieving an excellent accuracy, surpassing other classifiers like the complex support vector machine (CSVM). Another research conducted by Alamatsaz et al. [15] introduces a light deep learning approach for high accuracy in the classification of 8 different types of arrhythmias. The proposed method leverages resampling and baseline wander removal to prepare the dataset and utilizes a hybrid CNN-LSTM model for the classification. All of this literature demonstrated the practical application of different ML algorithms along with their effectiveness in cardiac arrhythmia classification.

This study involves the comprehensive exploration of two machine learning models for the classification of electrocardiogram (ECG) signals and the detection of arrhythmias. Starting with the exploration of morphological features of ECG signals obtained from the PhysioNet Challenge 2017 dataset [16], [17], a bidirectional LSTM model was developed using MATLAB, which distinguished between normal ECG signals and a prevalent form of arrhythmia, namely Atrial Fibrillation. The performance of other machine learning algorithms like decision trees, naïve Bayes classifiers and neural networks were further explored using MATLAB's machine learning toolkit. Utilizing the model's output in terms of accuracy, a questionnaire was developed and circulated to the medical community to collect their opinion and feedback on such systems. The survelogical result highly reinforced the potential value of such systems in assisting healthcare professionals and offered insightful suggestions for further improvement on the model's performance.

Progressing from the initial binary classification, a more robust model employing a 1D Convolutional Neural Network (1D-CNN) was developed for multiclass ECG signal classification. Training data was sourced from the MIT-BIH arrhythmia database [17], [18] and comprised five distinct types of ECG signals, including the normal beats, right bundle branch block beat (RBBB), left bundle branch block beat (LBBB), premature ventricular contraction (PVC) and premature atrial contractions (PAC). This Python-based model used a variety of libraries, including TensorFlow, Keras, SKlearn, Pandas, and Numpy [19]. A stratified 5-fold cross-validation technique [20] was employed to evaluate this model, achieving a promising accuracy across all folds. Moreover, other crucial metrics such as F1-score, precision, recall, and specificity [21] were calculated for each class, all indicating high values that further testify to the model's robustness and precision in classifying the five ECG signal types. Lastly, a web-based portal was developed locally using HTML and CSS for the real-time demonstration of the model's classification abilities using Python's Selenium web driver library to bridge the trained model and the interface.

Overall, this study explores the successful application of machine learning algorithms in classifying time-series signals, such as ECG, with an innovative approach of incorporating direct feedback from medical personnel into the system. This initiative led to valuable refinements, enhancing the model's performance and highlighting the practical applicability of these models in real-world healthcare settings.

## II. METHODOLOGY

The methodology employed in this research is illustrated in Figure [2].

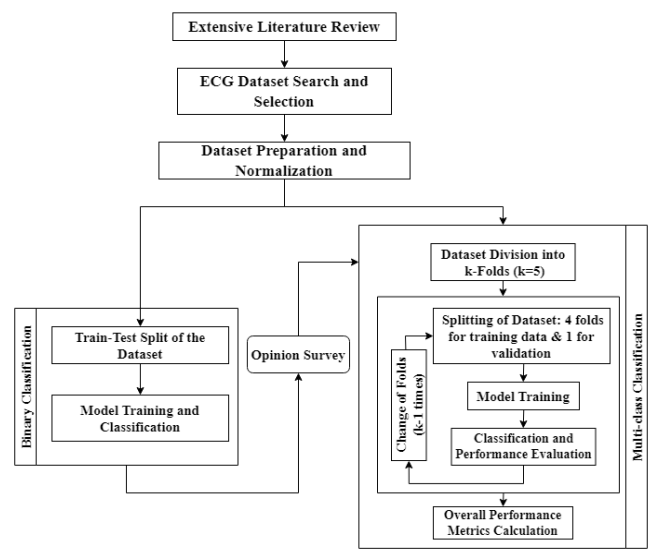

Fig. 2. Block diagram of methodology

For binary classification of ECG signals as normal or Atrial Fibrillation (Af), PhysioNet Challenge 2017 dataset from the PhysioNet website was utilized and converted into a MATLAB-compatible format. This dataset, comprising signals of various lengths and sampling frequencies, was standardized to a uniform length of 30 seconds. Wavelets transform was utilized to detect the QRS complex [22], allowing the detection and analysis of morphological features such as number of peaks, heart rate and peak-peak distance over a certain period. The data imbalance between normal and abnormal signals was mitigated through resampling [23], and additional features such as instantaneous frequency and spectral entropy were also extracted.

Prior to training the Bi-LSTM model, the dataset was normalized using Z-score normalization and split into training and testing sets in a 9:1 ratio. The choice of the Bi-LSTM model was informed by its established efficiency in processing time-series data [11], a key characteristic of ECG signals. This model, consisting of a sequence input layer, a bidirectional LSTM layer with 100 units, a fully connected layer, a SoftMax layer, and a classification layer, was trained using the Adam optimizer with specified parameters. Model performance was evaluated through accuracy and confusion matrix analysis for training and testing sets. Additionally, supervised machine learning models such as Decision Tree Classifier, Naïve Bayes classifier, and Neural Network were employed using MATLAB's machine learning toolbox for the comparative analysis.

A questionnaire was developed using the result from the binary classification, which primarily focused on collecting the opinion of the medical personnel on such systems and their suggestions to make the system more practical. It was then circulated among the medical community. The survey responses showcased an encouraging trend, with the majority of medical professionals expressing their readiness to incorporate AI-powered tools into their practice. Based on the confusion matrix presented, they acknowledged the potential of machine learning models in the detection of arrhythmias. The feedback from medical professionals uniformly supported the idea of broadening the model's capabilities by incorporating a few more types of arrhythmic classes.

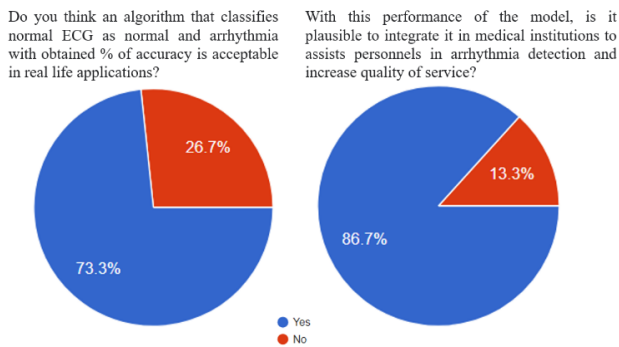

Fig. 3. Medical professional opinion survey results.

Figure [3] shows the chart showcasing the response of medical personnel to some crucial queries. The majority of responses were positive towards the query, but a certain percentage suggested the opposite. This was due to the issue of 'false negatives' shown in the confusion matrix, which showed the number of Abnormal signals classified as normal ones. This issue in the practical world may lead to the ignorance of some dreadful cardiac situations, resulting in extreme casualties. The need to improve the model's accuracy as much as possible was hence highlighted, which would, in turn, increase its effectiveness as a diagnostic tool.

Building upon the insights from the initial binary classification, a more sophisticated model using a 1-Dimensional Convolutional Neural Network (1D-CNN) was developed using Python. The chosen dataset was the MIT-BIH arrhythmia dataset from the PhysioNet website, encompassing five distinct types of ECG signals: Normal ECG signals, right bundle branch block beat (RBBB), left bundle branch block beat (LBBB), premature ventricular contraction (PVC), and premature atrial contractions (PAC). Using 'sym4' wavelets, maximal PQRST waveforms and their respective annotations were extracted. The dataset was then normalized using z-score normalization. Given the substantial class imbalance in the dataset, with a predominance of normal beats, resampling techniques were employed to balance the distribution. A stratified 5-fold cross-validation strategy was implemented for splitting the dataset. Notably, a subset of signals from each class was separately preserved for real-time demonstration in the web portal. These processing steps leverage Python's capabilities, using libraries such as Tensorflow, Keras, SKlearn, Pandas, NumPy, and Matplotlib.

The model architecture consisted of a 1-Dimensional Convolutional Neural Network structured into three blocks, each consisting of a convolutional layer, a batch normalization layer and a pooling layer. With kernel sizes of 7, 5, and 3, and filters of 32, 64, and 128, respectively, hierarchical feature extraction from data was facilitated. The output was flattened and forwarded to fully connected layers consisting of Dropout layers for L2 regularization to reduce overfitting. These fully connected layers comprised 128 and 64 neurons and utilize the ReLU activation function. The output layer employed a softmax activation for class probabilities, with the model optimized using Adam and evaluated via accuracy.

As a stratified 5-fold cross-validation technique was employed, the model was trained five times separately, rotating the folds to generate different training and testing data sets, ensuring that each fold once served as the testing set. This ensured that every data on the dataset was used for both training and testing purposes, resulting in a better evaluation of the model's performance.

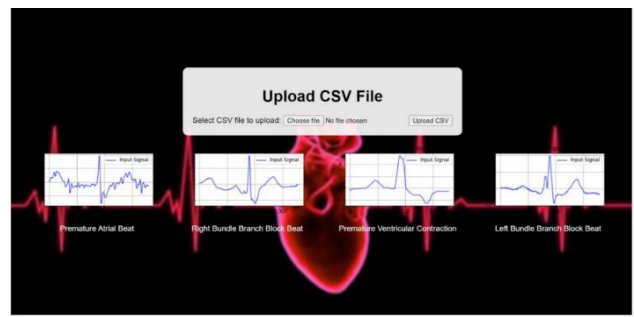

Fig. 4. Screenshot of the web-based portal

In order to showcase the real-time classification capabilities of the model, a web-based portal was developed utilizing HTML and CSS, as shown in Figure [4]. The Selenium Webdriver served as a bridge, linking the user interface with the trained model. Importantly, this setup was deployed on a local host for immediate and practical demonstrations.

### III. Training and Testing Dataset

Two different datasets, PhysioNet Challenge 2017 and MIT-BIH arrhythmia, were considered for binary classification using MATLAB and multilevel classification using CNN, respectively. The PhysioNet Challenge 2017 contains 8528 ECG recordings, sampled at 300 Hz, mostly 30 seconds long and some were up to 60 seconds, containing each signal classified as either normal rhythm or atrial fibrillation, one of the most common arrhythmias. It can be identified by the missing P block in the PQRST waveform and often causes irregular heartbeat.

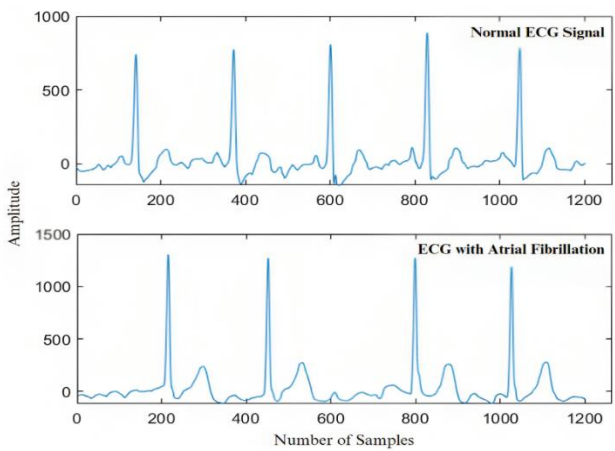

Fig. 5. Normal ECG and ECG with atrial fibrillation.

There was also the presence of a notably very small number of other noisy signals that were not classified, and thus, were removed from the dataset, resulting in a decrease in the total number of signals. After truncating long signals to make it 30 seconds, 5655 total numbers with a 7:1 normal to abnormal ratio were obtained. A resampling technique was used to balance this ratio in the dataset. The 'sym4' wavelet was used for QRS detection and artefact removal, enabling the detection of morphological features like R-R intervals, distance between R peaks and heart rate. Other features like spectral entropy and instantaneous frequency, which were affected by the missing P waves, were explored using MATLAB's signal processing toolbox. The prepared dataset was split into 9:1 training to testing data ratio.

The 1D CNN model was considered for multilevel classification and the MIT-BIH arrhythmia dataset was obtained from the PhysioNet website [16]. The MIT-BIH arrhythmia database directory is a set of over 4,000 long-term recordings from a medical instrument called Holter obtained in the Beth Israel Hospital (BIH) arrhythmia laboratory. The samples were obtained from 25 men and 22 women aged 23 to 89. These readings were taken from two leads, lead II and lead VI, of a 12-lead ECG system. The signals were digitized with a sampling frequency of 360 Hz. After pulling the dataset, the PQRST wave was extracted by comparing the beats with 'sym4' signals and using wavelet coefficients. All these signals and their respective annotation were stored as a data frame. The data frame consisted of 5 types of signals annotated as N (Normal beat), L (Left bundle branch block beat), A (Atrial premature beat), R (Right bundle branch block beat), and V (Premature ventricular contraction) having a total sum of 100012 beats.

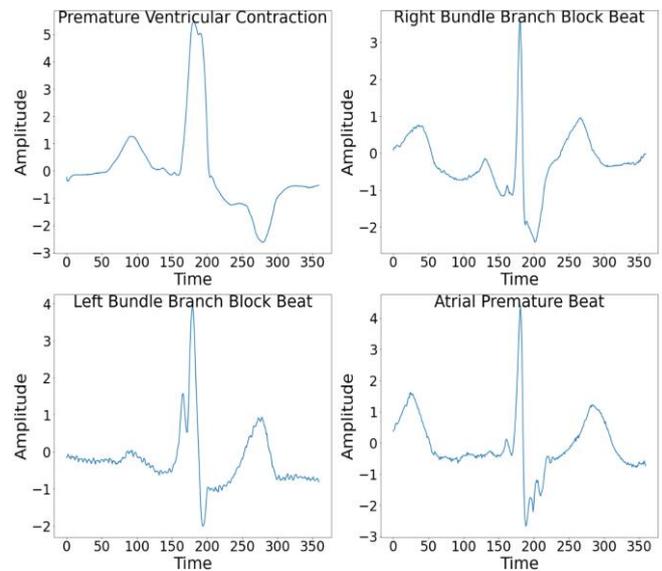

Fig. 6. ECG waveforms for various types of arrhythmias.

Figure [6] shows the four types of arrhythmias present in the signals available in the dataset. L (Left Bundle Branch Block Beat) is characterized by the disruption of the electrical signal travelling through the left bundle branch, causing the left ventricle to contract later than it should, leading to an aberration in the rhythm pattern. A (Atrial Premature Beat) is an early initiation of an electrical signal in the atria, which leads to premature contraction and can alter the regular rhythmic cycle. R (Right Bundle Branch Block Beat) refers to a delay along the pathway of the electrical signal as it passes through the right bundle branch, resulting in asynchronous contractions of the ventricles. V (Premature Ventricular Contraction involves an early and abnormal contraction originating in the ventricles, often leading to the sensation of a "skipped beat".

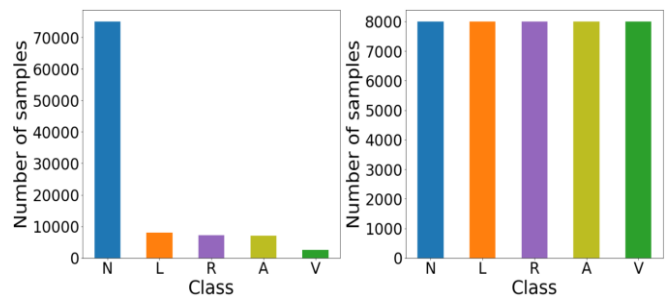

Fig. 7. Dataset before and after resampling.

It can be observed in Figure [7] that in the MIT-BIH dataset, there was a massive imbalance between the number of beats of each class in the dataset among various classes of data. The resampling technique of feature engineering was applied to balance the dataset, which resulted in the number of beats of each class to 8000.

A stratified k-fold cross-validation technique was employed with k set to 5. This process ensures that each fold represents the whole dataset well, making it suitable for an imbalanced dataset. During each iteration out of 5, 4 folds (or 80%) of the dataset were used to train the model, and the remaining fold (or 20%) was used for validation. This dataset-splitting process was repeated five times, with each fold acting as the validation set precisely once. This way, it was ensured that every sample in the dataset was used for both training and testing, generating a robust model performance.

## IV. RESULTS AND DISCUSSIONS

### A. Arrhythmia Detection Performance using Bi-LSTM

Exploration and analysis of the morphological features of ECG signals, such as peak-to-peak distance, heart rates and number of QRS complexes over a window of samples, visually distinguished the normal ECG signals and the abnormal ones (the ones with atrial fibrillation). In the window of 1200 samples, the number of QRS complexes observed was hovering between 4 and 6 in the normal signals, which yielded a heart rate between 60 bps and 90 bps. The peak-to-peak distances were relatively similar in normal signals but anomalous in atrial fibrillation. Heart rate and number of QRS complexes in the atrial fibrillation signals significantly deviated from those values of normal signals. Two more features, instantaneous frequency and spectral entropy, were extracted using MATLAB's signal processing toolbox. After training the build Bi-LSTM model, training and testing accuracy of 94.64% and 92.44% were obtained respectively.

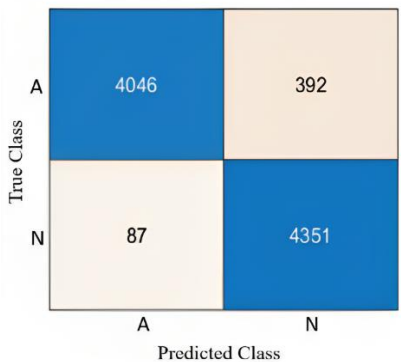

Fig. 8. Bi-LSTM model confusion matrix.

Figure [8] shows the confusion matrix plotted after feeding the test data into the trained Bi-LSTM model. Upon training other models like the Decision Tree Classifier, Naïve Bayes Classifier and Neural Network using MATLAB's deep learning toolbox, relatively low testing accuracies such as 75.3%, 71.3%, and 73.3%, respectively, were obtained. This verified the better performance of the Bi-LSTM model in the classification of time series data such as ECG.

### B. Arrhythmia Classification using CNN

The proposed 1D-CNN model was evaluated using a stratified 5-fold cross-validation technique, where each fold was trained independently using 36 batches for 50 epochs. This strategy balanced the representation of each class across all the folds, enhancing the model's performance. The model exhibited a high degree of stability in all folds, with a maximum accuracy of 99.24% and minimum accuracy of 98.66%. A higher value of minimum accuracy indicates the model's robustness and reliable performance.

In evaluating model performance, confusion matrices from all the folds were combined to create a consolidated confusion matrix shown in Figure [9].

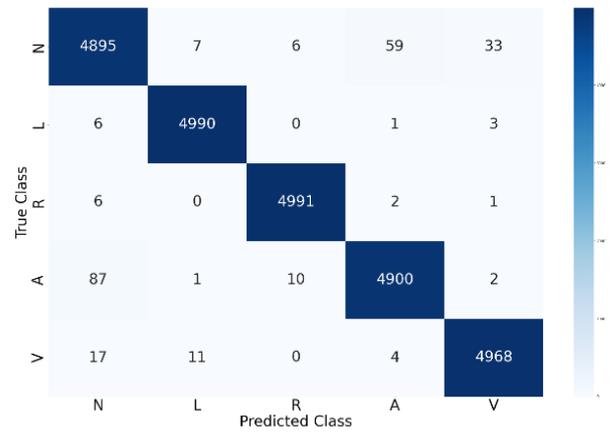

Fig. 9. Multiclass classification confusion matrix.

The collective confusion matrix of all folds shows that the model had an overall strong performance across all classes. However, to have a comprehensive view of the model's performance, a table with additional metrics like Precision, Recall, F1-score, Specificity, and class-wise accuracy were calculated, shown in Table [1].

TABLE I. CLASSIFICATION MODEL PERFORMANCE PARAMETERS

| Class | Precision | Recall | F1-Score | Specificity | Accuracy |
|---|---|---|---|---|---|
| N | 0.9768 | 0.9790 | 0.9779 | 0.9942 | 0.9790 |
| L | 0.9962 | 0.9980 | 0.9971 | 0.9990 | 0.9980 |
| R | 0.9968 | 0.9982 | 0.9975 | 0.9992 | 0.9982 |
| A | 0.9867 | 0.9800 | 0.9933 | 0.9967 | 0.9800 |
| V | 0.9922 | 0.9936 | 0.9929 | 0.9980 | 0.9936 |

From the detailed evaluation Table [1], it is evident that the model had high precision, recall or sensitivity and F1-score across all the classes, indicating successful recognition of the different arrhythmias. These results demonstrate the effectiveness of the model in handling multi-class classification, highlighting its potential use in real-world applications for computer-aided arrhythmia diagnosis.

### C. Scope and Limitations

The results of this study, along with the medical expert's opinions persuade a transformative era in the healthcare industry. It promises enhanced clinical diagnostics, paving the way for precise and timely interventions. The potential extends to telemedicine, assistive diagnosis, educational and training tools, interdisciplinary collaboration between cardiologists and data scientists, AI-driven wearables and much more, leading to a seamless blend of modern and innovative health practices with existing infrastructures.

On the other hand, the physiology of the human body can vary based on geographic location, genetics, environmental factors, and other determinants. This variation raises the challenge of gathering comprehensive datasets to train machine learning models effectively. A model trained on data from one specific region might not perform optimally in a different setting. For instance, the unique physiological attributes found in the Nepalese population might differ from those in the dataset. Additionally, while deep learning models offer precision, their computational demands might not be feasible for real-time applications in resource-constrained settings. Transitioning from traditional healthcare systems to AI-based diagnostics, particularly in regions like Nepal, presents its own set of challenges. Integrating sophisticated AI models into the existing infrastructure requires not only technological adaptations but also training, policy adjustments, and cultural acceptance. Making such a transition is far from straightforward.

V. SUMMARY AND CONCLUSIONS

The integration of machine learning systems in interactive healthcare services has the potential to revolutionize the industry and enhance medical diagnostic capabilities. Medical professionals' express enthusiasm for implementing such models, signaling a promising future for AI-driven healthcare solutions.

This paper is focused on exploring and developing multiple machine learning models for electrocardiogram (ECG) signal classification to facilitate the detection and classification of arrhythmias. The investigation showed that the Bi-LSTM model outperformed other conventional models like Decision Tree Classifier, Naïve Bayes Classifier and Neural Network in binary classification. Bi-LSTM could achieve remarkably high accuracy in binary classification or arrhythmia detection. After identifying and visualizing the morphological features effectively, the initial Bi-LSTM achieved satisfactory accuracy in binary classification. The opinion survey of medical professionals on the performance and applicability of ML-based detection and classification systems following the initial results supported the practicality of this kind of system. Meanwhile, the experts highlighted the traits and provided feedback to improve the system's performance. Following the suggestions, a more robust model, CNN, was used for multiclass-level classification of arrhythmias. This stratified-5-fold cross-validation employed model yielded highly acceptable results in accuracy, precision, recall, and f-measure for classifying the normal ECG signals and four types of arrhythmias. This tool can potentially become a valuable diagnostic asset for medical personnel with further enhancements.

In conclusion, the successful development and validation of ML models for ECG signal classification underscore the potential for the widespread adoption of such technologies in the healthcare industry. As the integration of artificial intelligence advances, more sophisticated models in computer-aided diagnostic (CAD) systems can be anticipated.